\documentclass[conference]{IEEEtran}

\usepackage[cmex10]{amsmath}
\usepackage{amsthm}
\usepackage{amssymb}
\usepackage{mathrsfs}
\usepackage{graphicx}
\usepackage{float}
\usepackage{array}
\usepackage{epstopdf}
\usepackage{multirow}
\usepackage{cite}

\begin{document}

\title{\huge{I\MakeLowercase{ris}-GAN:
L\MakeLowercase{earning}
\MakeLowercase{to}
G\MakeLowercase{enerate} R\MakeLowercase{ealistic} I\MakeLowercase{ris} 
I\MakeLowercase{mages}
U\MakeLowercase{sing}
C\MakeLowercase{onvolutional}
GAN  }}

\author{Shervin Minaee$^*$, AmirAli Abdolrashidi$^{\dagger}$  \\
$^*$New York University
\\ $^{\dagger}$University of Californa, Riverside\\ \\
}

\maketitle

\begin{abstract}
Generating iris images which look realistic is both an interesting and challenging problem. Most of the classical statistical models are not powerful enough to capture the complicated texture representation in iris images, and therefore fail to generate iris images which look realistic.
In this work, we present a machine learning framework based on generative adversarial network (GAN), which is able to generate iris images sampled from a prior distribution (learned from a set of training images).
We apply this framework to two popular iris databases, and generate images which look very realistic, and similar to the image distribution in those databases.
Through experimental results, we show that the generated iris images have a good diversity, and are able to capture different part of the prior distribution.
We also provide the Frechet Inception distance (FID) obtained for the generated iris images by our model, and show that our model is able to achieve very promising FID score.
\end{abstract}

\IEEEpeerreviewmaketitle

\section{Introduction}
\label{sec:Intro}
Iris, along with face \cite{face_intro}, fingerprint \cite{finger_intro}, and palmprint \cite{palm_intro}, is among popular biometrics used for identity recognition and security applications. When it comes to recognition tasks, Some of the advantages iris possesses are that it can easily be sampled, it contains many rich features, and it is mostly protected from environmental factors and changing due to aging \cite{daugman_howirisworks}. 
Iris recognition is widely used for security applications such as national border controls, airports, cellphone authentication.

There have been numerous approaches proposed for iris recognition. 
One of the earliest is the work by Daugman \cite{daugman_1st} which used 2D Gabor wavelet transform for this task. 
There are many other approaches based on hand-crafted features, such as Haarlet pyramid features \cite{iris-haar}, scattering features \cite{iris-scat}, and random projection based features \cite{iris-random}.
With the rise of deep neural networks in recent years \cite{cnn1,cnn2,cnn3}, there have been even more methods proposed for iris and biometrics recognition in general  \cite{irisdeep1,irisdeep2, fingerdeep, palmdeep}. 

All biometrics, including iris, have unique patterns. Therefore, generating an iris image would not be easy at all. However, a new trend has begun in recent years using a variety of different methods. Wei et al \cite{iris_gen1} managed to generate iris images using patch-based sampling. Makthal \cite{iris_gen2} proposed to use Markov random fields (MRF) to generate synthetic iris images. Cui \cite{iris_gen3} first synthesizes iris images using principal component analysis (PCA) and then uses super-resolution to enhance them. Zuo \cite{iris_gen4} uses a ``model-based anatomy-based'' approach, which includes generating 3D fibers in a cylindrical shape and projecting them on a 2D field, followed by adding blur effects and eyelids.

Our work mainly is based on generative adversarial networks (GAN) \cite{gan}, which provide a powerful framework for learning to generate data samples from a given distribution.  
A GAN is comprised of a \textit{generative network}, which learns a training dataset's distribution and can generate new data (sampled from the same distribution), and a \textit{discriminative network}, which tries to distinguish real samples from the ones generated with the generator model. 
Since their invention, GANs have been used for various image generation tasks (such as digits, bedroom images, house numbers, faces, etc.), as well as other applications, including image super-resolution \cite{sr-gan}, generating patterns in music \cite{gan_music}, and also in fashion \cite{gan_fashion}.

With this in mind, it can be deduced that biometric images can now potentially be generated and used for various purposes, including anti-spoofing applications. 

\begin{figure}[h]
\begin{center}
   \includegraphics[width=0.9\linewidth]{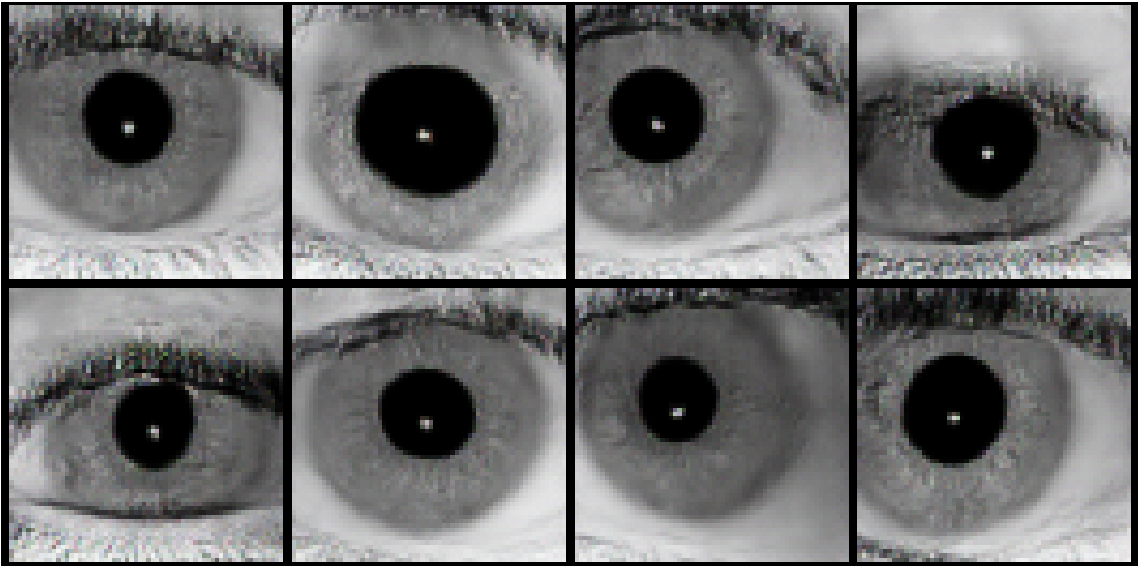}
\end{center}
   \caption{Eight sample generated iris images using the proposed framework. These images are generated with a model trained on IIT Delhi Iris database.}
\label{fig:gan_model}
\end{figure}

In this work, we propose an iris image generation framework based on deep convolutional generative adversarial networks (DC-GAN), which is able to generate realistic iris images, which cannot be distinguished from real ones.
Figure \ref{fig:gan_model} provide 8 sample images generated by our model.
As we can see, the generated images with our model have iris, cornea, pupil (with different zooming), eyelashes, etc. similar to real iris images.

As mentioned earlier, generative adversarial networks \cite{gan} provide a powerful framework for learning to generate data samples from a given distribution. 
On high level, they simultaneously train a generator and discriminator model, which the prior one learns to generate samples similar to real data, and the later one tries to discriminate real samples from ``fake'', i.e. the ones generated with the generator.
Since the invention of GAN, they have been widely used for various applications, and several extended version of the vanilla GAN model are proposed. 
We will discuss more about some of the extensions of GANs in the next section.

The structure of the rest of this paper is as follows.
Section \ref{sec:Framework} provides the details of our proposed framework, and the model architectures for both the generator, and discriminator networks.
In Section \ref{sec:Evaluation}, we provide the experimental studies, and present the generated iris images with our proposed framework. We show the generated iris images over different epochs (for the sample input noise) to see how the generated images evolve over time.
And finally the paper is concluded in Section \ref{sec:Conclusion}.

\section{The Proposed Framework}
\label{sec:Framework}
In this work we propose a generative model for iris images based on deep convolutional generative adversarial networks. 
To provide more details about how GANs work, the generator network in GANs learns a mapping from noise $z$ (with a prior distribution) to a target distribution $y$, $G= z \rightarrow y$.
The generator model, $G$, tries to generate samples which look similar to the ``real'' samples (provided during training), while the discriminator network, $D$, tries to distinguish the samples generated by the generator models from the ``real'' ones.

\begin{figure}[h]
\begin{center}
   \includegraphics[page=1,width=0.98\linewidth]{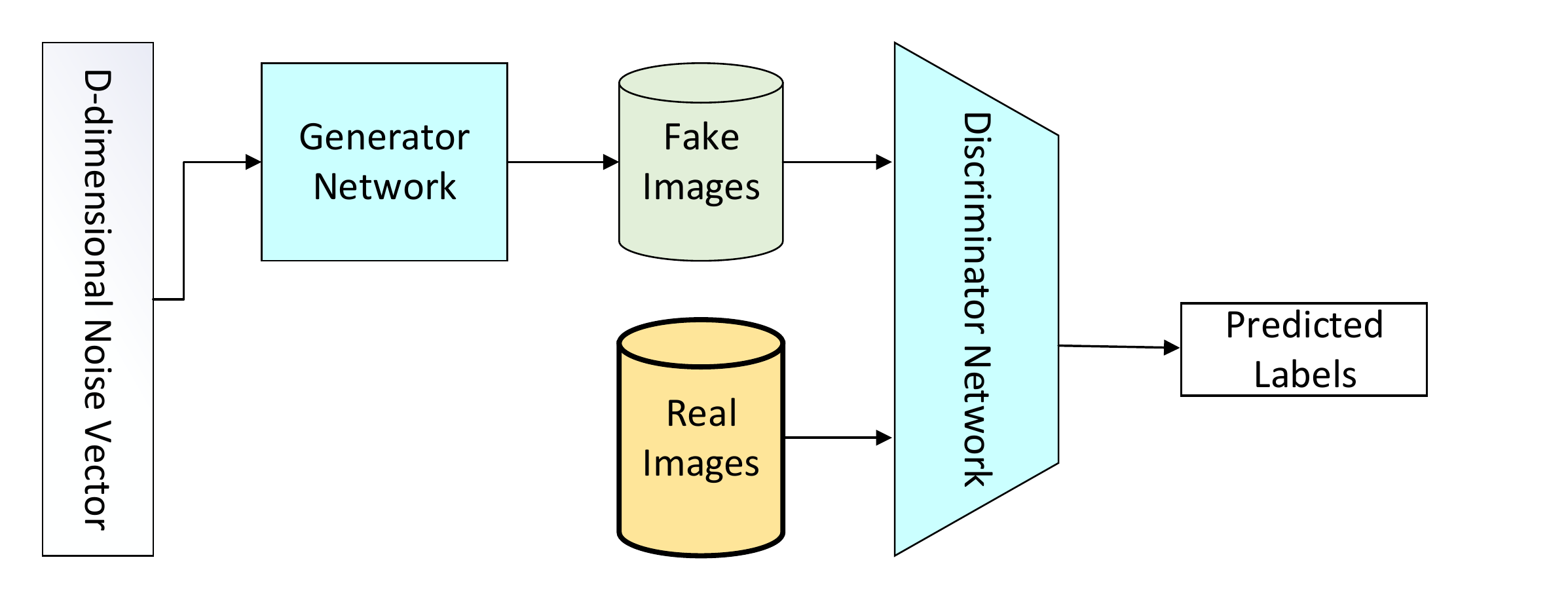}
\end{center}
   \caption{The general architecture of generative adversarial networks}
\label{fig:gen_arch}
\end{figure}

The general architecture of a GAN model is shown in Figure \ref{fig:gen_arch}.

The loss function of GAN can be written as:
\begin{equation}
\begin{aligned}
& \mathcal{L}_{GAN}=  \mathbb{E}_{x \sim p_{data}(x)}[\text{log} D(x)]+ \mathbb{E}_{z \sim p_z(z)}[\text{log}(1-D(G(z)))]
\end{aligned}
\end{equation}
In which we can think of GAN, as minimax game between $D$ and $G$, where $D$ is trying to minimize its classification error in detecting fake samples from real ones (maximize the above loss function), and $G$ is trying to maximize the discriminator network's error (minimize the above loss function).
After training this model, the trained generator model would be: 
\begin{equation}
\begin{aligned}
G^*= \text{arg} \ \text{min}_G \text{max}_D \ \mathcal{L}_{GAN}
\end{aligned}
\end{equation}
In practice, the loss function in Eq (1) above may not provide enough gradient for $G$ to get trained well, specially at the beginning where $D$ can easily detect fake samples from the real ones. 
One solution is to instead of minimizing $\mathbb{E}_{z \sim p_z(z)}[\text{log}(1-D(G(z)))]$, one can train it to maximize $\mathbb{E}_{z \sim p_z(z)}[\text{log}(D(G(z)))]$.


Since invention of GAN, there have been several works trying to improve/modify GAN in different aspect.
To name a few promising works, in \cite{dc-gan}, Radford and authors proposed a convolutional GAN model for generating images, which works better than fully connected networks when used for image generation.
In \cite{con-gan}, Mirza proposed a conditional GAN model, which can generate images conditioned on class labels. This is a nice modification of the vanilla GAN model, which enables one to generate samples with a specified label.
The original paper on GAN uses KL divergence to measure the distance between distributions, and a problem can happen if these distribution have non-overlapping support in which KL divergence is not a good representative of their distance. 
The work in \cite{was-gan} tries to address this issue, by proposing a new loss function based on Wasserstein distance (also known as earth mover's distance).
In \cite{cycle-gan}, Zhu proposed an image to image translation model based on a cycle-consistent GAN model, which learns to map a given image distribution into a target domain (e.g. day to night images, summer to winter images, horse to zebra, etc).
In \cite{sr-gan}, Ledig proposes an image super-resolution approach based on GAN, which tries to generate high-resolution version of images which look similar to the target high-resolution images.
The idea of adversarial training has been also applied to auto-encoder framework, to provide an unsupervised feature learning approach \cite{adv_ae1, adv_ae2} (here the adversarial module is trying to distinguish samples for which the latent representation is coming from a prior distribution from other samples).
There are many other works extending GAN models in different ways. 
For a detailed list of works relevant to GAN, please refer to \cite{gan-zoo}.

\begin{figure}[h]
\begin{center}
   \includegraphics[page=2,width=0.98\linewidth]{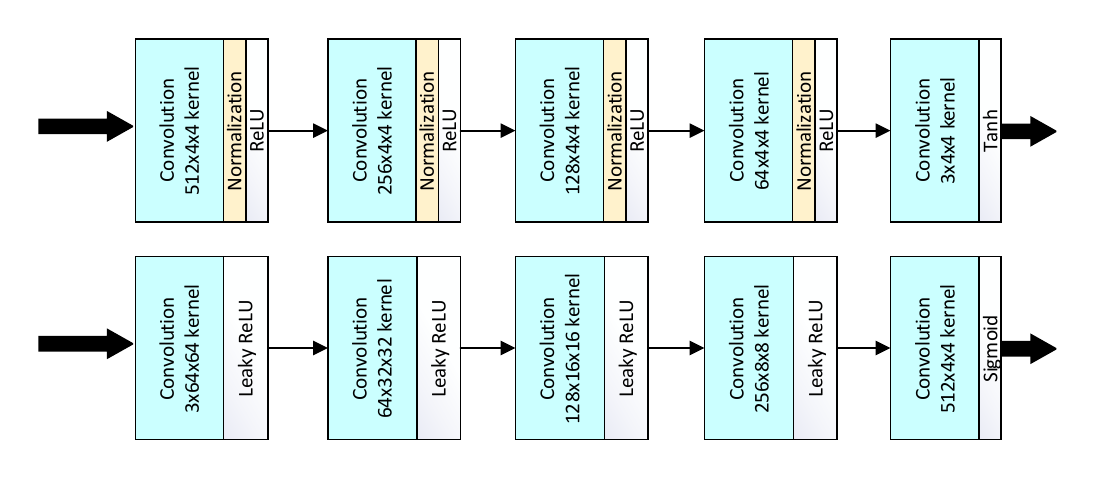}
\end{center}
   \caption{The architecture of the proposed generator (top) and discriminator model (bottom) by our work.} 
\label{fig:our_model}
\end{figure}

In our work, we use a simple DC-GAN model for learning to generate iris images. 
In our framework, we use simple convolutional networks for both the generator and discriminator models.
The discriminator model consist of 5 convolutional layers followed by batch-normalization and leaky ReLu as non-linearity.
The generator model contains 5 fractionally strided convolutional layers (A.K.A. deconvolution in some literature), followed by batch-normalization and non-linearity.
The architecture of our generator and discriminator networks are shown in Figure \ref{fig:our_model}.
It is really interesting that we are able to generate realistic iris images (as shown in experimental results) with these simple networks.

\section{Experimental Results}
\label{sec:Evaluation}
Before presenting the generated images by the proposed model, let us first discuss the hyper-parameters used in our training procedure.
We train the proposed model for 120 and 140 epochs (on CASIA and IIT Delhi databases respectively) using a Nvidia Tesla GPU.
The batch size is set to 80, and ADAM optimizer is used to optimize the loss function, with a learning rate of 0.0002. 
We also tried decaying the learning rate, but it did not change the final results significantly.
The input noise to the generator network, is 100 dimentional Guassian distribution with zero mean, unit variance.
We present the details of the datasets used for our work, and also the experimental results in the following subsections.

\subsection{Databases}
In this section, we provide a quick overview of two popular iris databases used in our work, CASIA-Iris dataset \cite{casia1000}, and IIT Delhi Iris dataset \cite{iit_paper}.

\begin{figure}[h]
\begin{center}
   \includegraphics[width=0.92\linewidth]{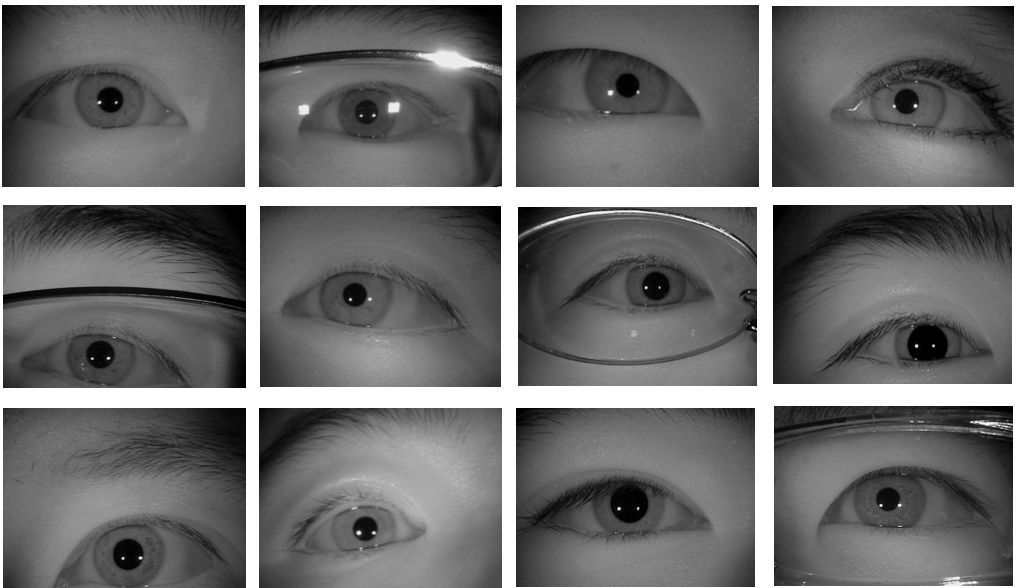}
\end{center}
   \caption{Twelve sample images from CASIA-1000 Iris database.}
\label{fig:casia}
\end{figure}

\textbf{CASIA Iris Dataset}:
CASIA has provided multiple iris databases. In this work, we use CASIA-Iris-Thousand dataset \cite{casia1000}, which contains 20,000 iris images from 1,000 subjects, which were collected using IKEMB-100 camera \cite{casia_site}.
The main sources of intra-class variations in CASIA-Iris-Thousand are eyeglasses and specular reflections.
Twelve sample images from this database are shown in Figure \ref{fig:casia}.

\begin{figure}[h]
\begin{center}
   \includegraphics[width=0.92\linewidth]{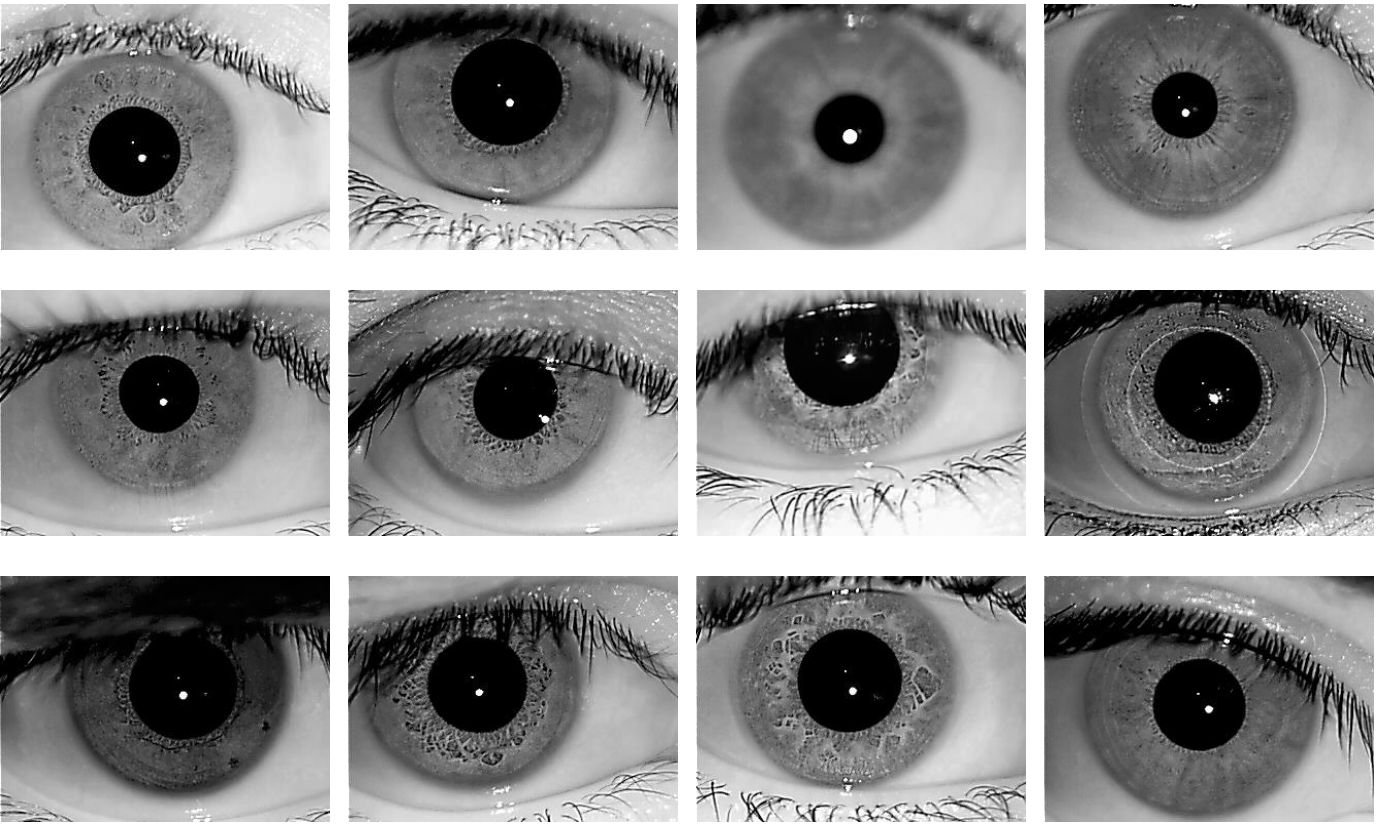}
\end{center}
   \caption{12 sample images from IIT Delhi Iris database.}
\label{fig:delhi12}
\end{figure}

\textbf{Delhi Iris Dataset}:
The IIT Delhi Iris Database consists of iris images collected from the students/staff at IIT Delhi  \cite{iit_paper}-\cite{iit_site}. 
The currently available database is from 224 users.
All the subjects in the database are in the age group 14-55 years comprising of 176 males and 48 females. 
The resolution of these images is 320x240 pixels and all these images were acquired in the indoor environment.
Twelve sample images from this database are shown in Figure \ref{fig:delhi12}.

\subsection{Experimental Analysis and The Generated Iris Images}
We will now present the experimental results of the proposed model on the above datasets.
We trained one model per dataset in our experiments, with the same model parameters discussed above.
For CASIA model, we only used images from a subset of 200 subjects to train the DC-GAN model, whereas for IIT database we used images from all 224 subjects.
After training these models, we can generate different images, by sampling latent representation from our prior Gaussian distribution and feeding them to the generator network of the trained models.

\begin{figure}[h]
\begin{center}
    \begin{tabular}{cc}
    \rotatebox{0}{0th Epoch} &
    \rotatebox{0}{20th Epoch} \\
    \includegraphics[width=0.45\linewidth]{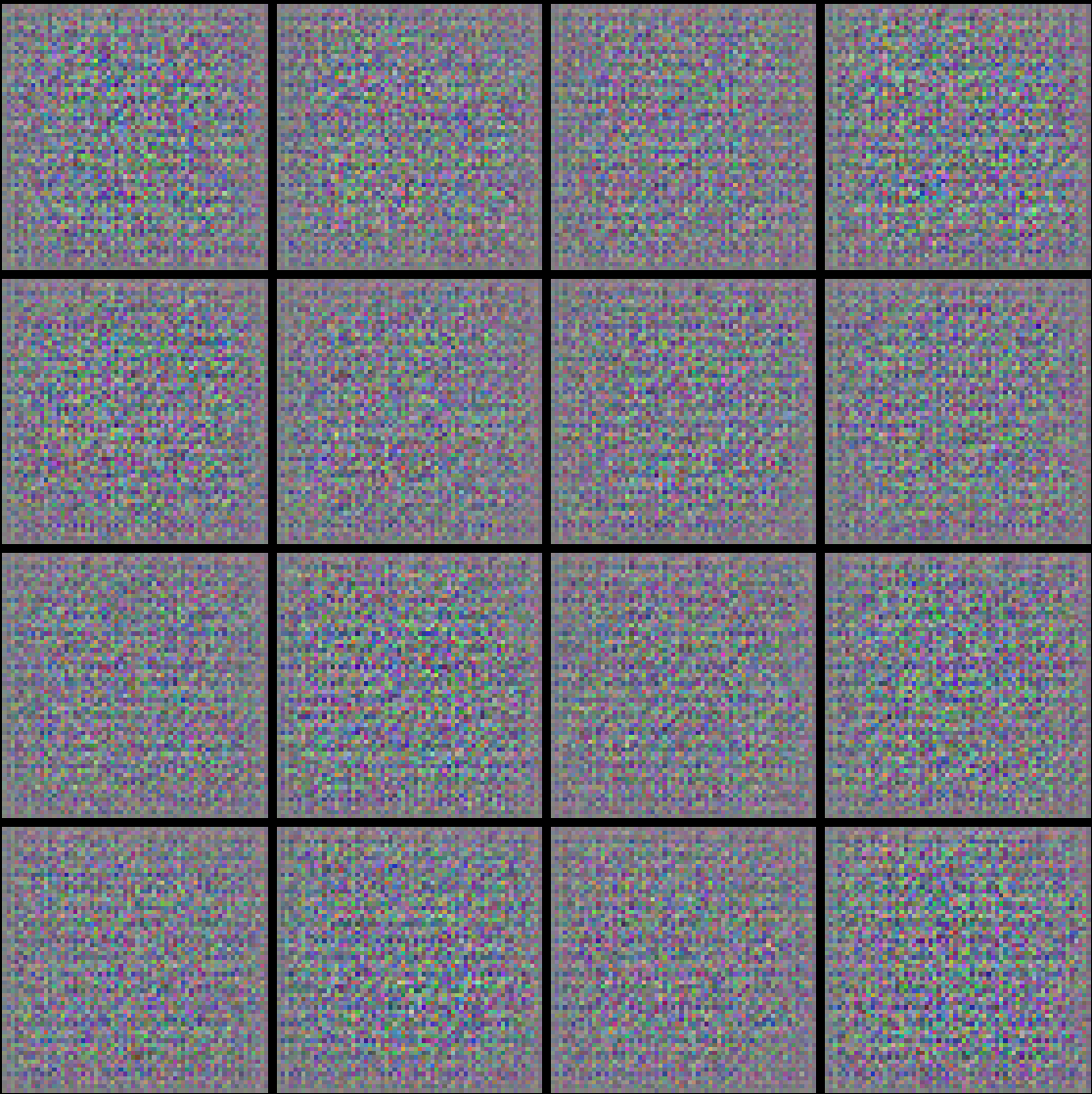}
    & \includegraphics[width=0.45\linewidth]{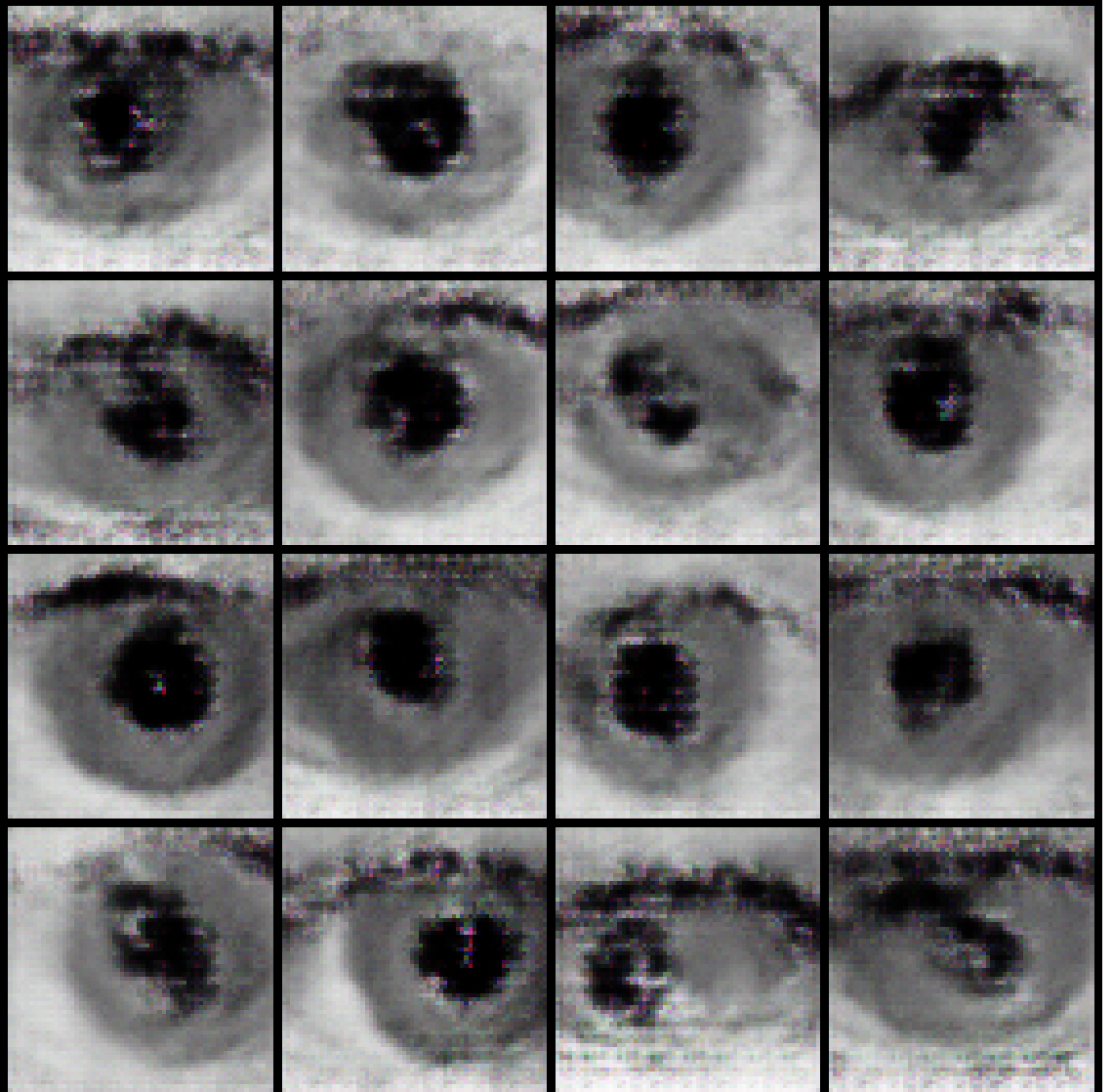}\\ 
    \rotatebox{0}{40th Epoch} &
    \rotatebox{0}{60th Epoch} \\
    \includegraphics[width=0.45\linewidth]{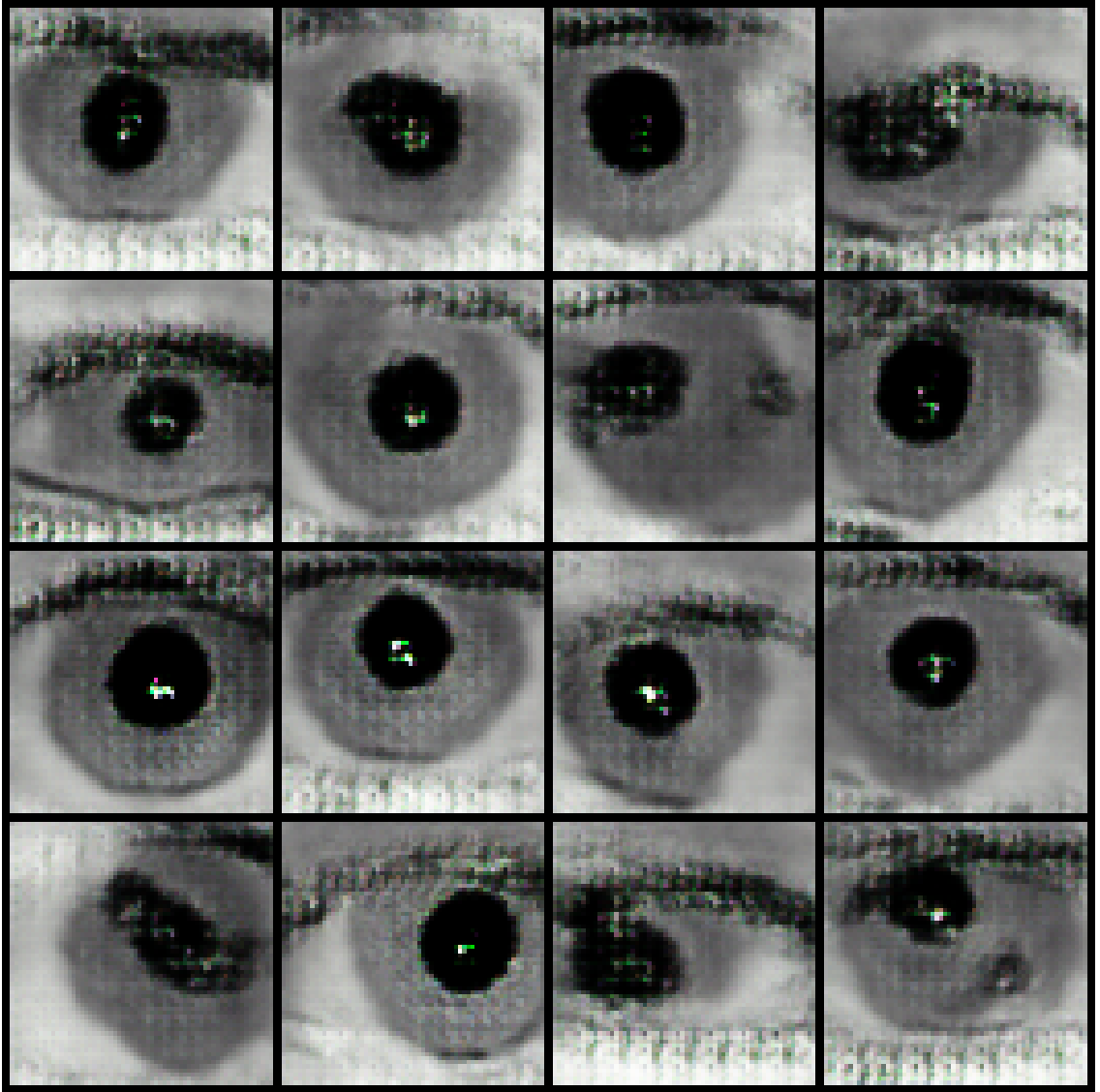}
    & \includegraphics[width=0.45\linewidth]{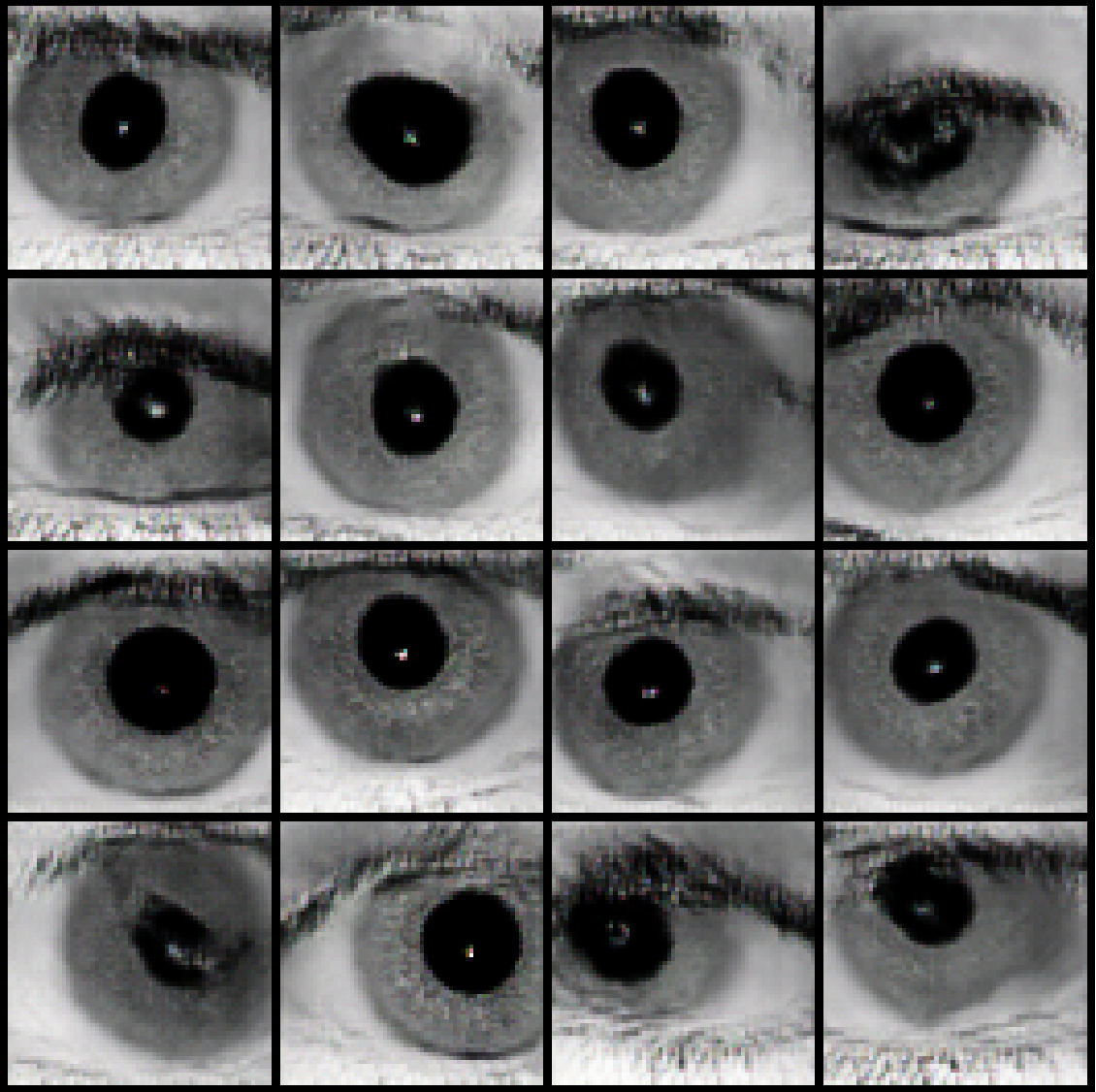}\\ 
    \rotatebox{0}{80th Epoch} &
    \rotatebox{0}{100th Epoch} \\
    \includegraphics[width=0.45\linewidth]{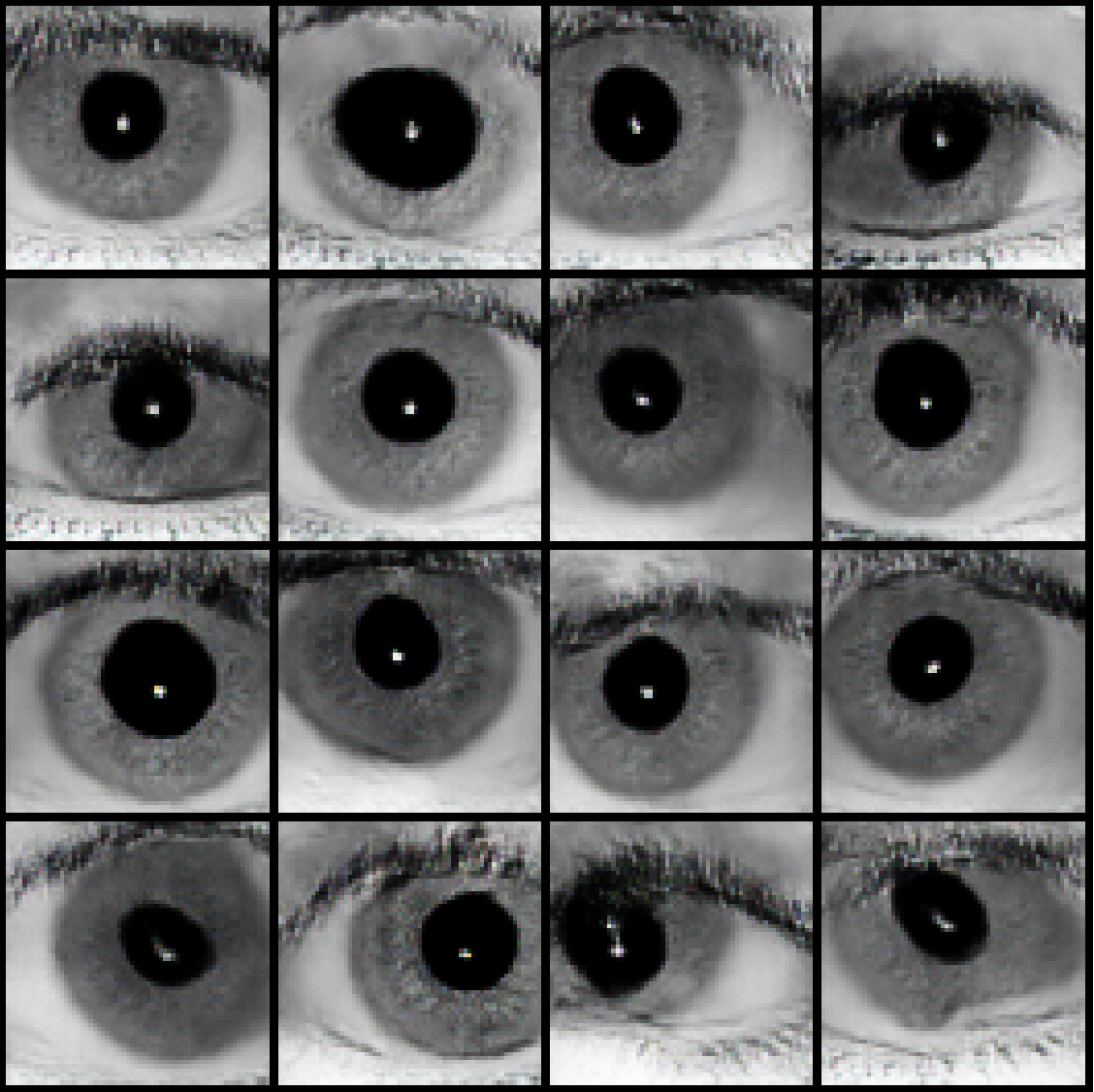}
    & \includegraphics[width=0.45\linewidth]{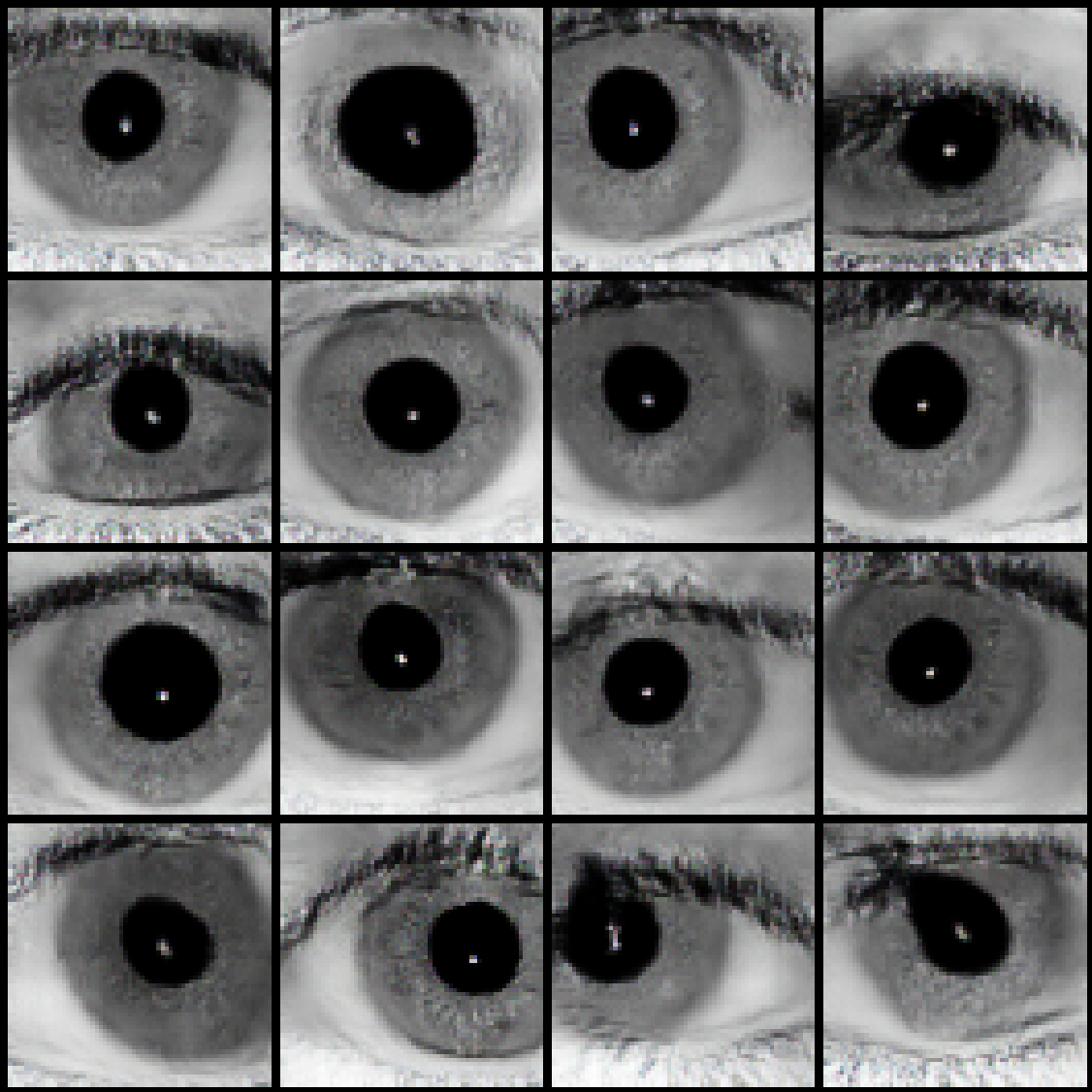}\\ 
    \end{tabular}
\end{center}
   \caption{16 generated iris images using the same input noise, over different epochs.}
\label{fig:CASIA16}
\end{figure}

\begin{figure*}[t]
\begin{center}
   \includegraphics[width=0.99\linewidth]{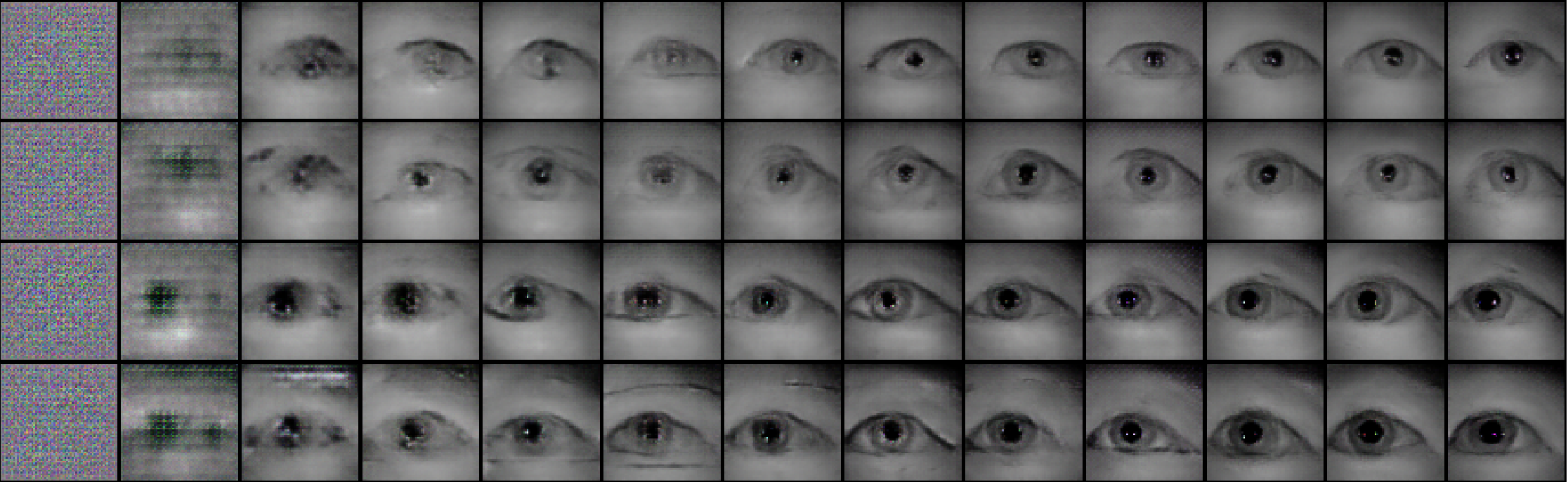}
\end{center}
   \caption{The generated iris images for 4 input latent vectors, over 120 epochs (on every 10 epochs), using the trained model on CASIA Iris database.}
\label{fig:casia_generate}
\end{figure*}

In Figure \ref{fig:casia_generate}, we show the generated iris images for four different latent representations, over every tenth epochs, when the model is trained on CASIA Iris dataset.
This will provide more granular changes happen after training the model for 10 more epochs.
As we can see, the generative model gets better and better over time, generating more realistic iris images.

\begin{figure*}[t]
\begin{center}
   \includegraphics[width=0.99\linewidth]{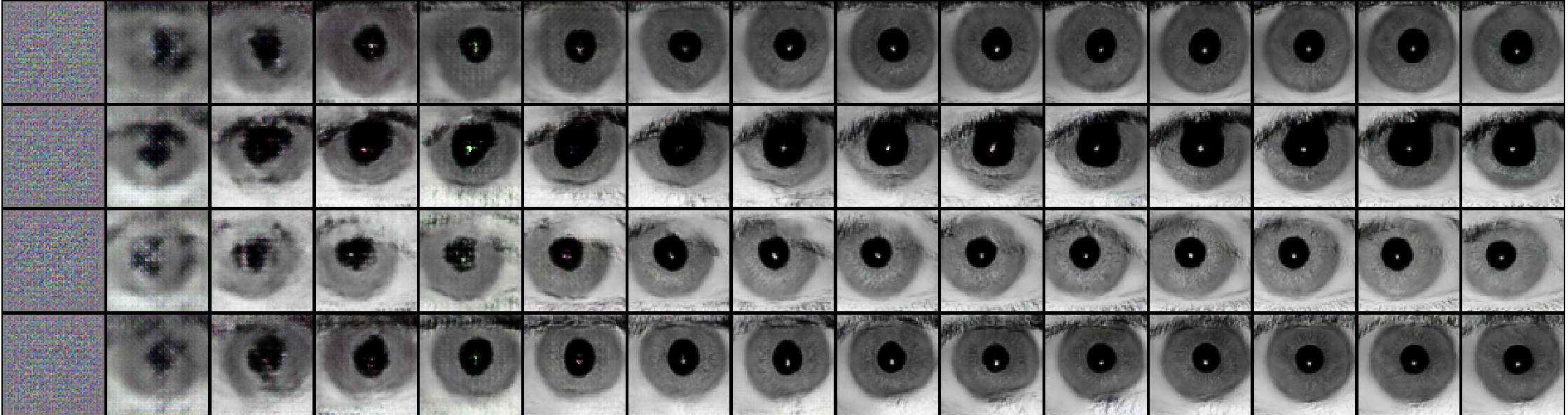}
\end{center}
   \caption{The generated iris images for 4 input latent vectors, over 140 epochs (on every 10 epochs), using the trained model on IIT Delhi Iris database.}
\label{fig:iitd_generate}
\end{figure*}

In Figure \ref{fig:iitd_generate}, we show a similar results generating four samples images, using the trained model on IIT Delhi Iris dataset.
We can make the same observation as above, where the generated sample images gets more realistic over time.

In Figure \ref{fig:CASIA16}, we present 16 generated iris images over different epochs (0th, 20th, 40th, 60th, 80th and 100th),  to see the amoung of diversity among the generated images.
As it can be seen, there is a good amount of diversity (in terms of the position of eyelashes, size of pupil, etc.) across the 16 generated images in this case.

\begin{figure}[h]
\begin{center}
   \includegraphics[width=0.95\linewidth]{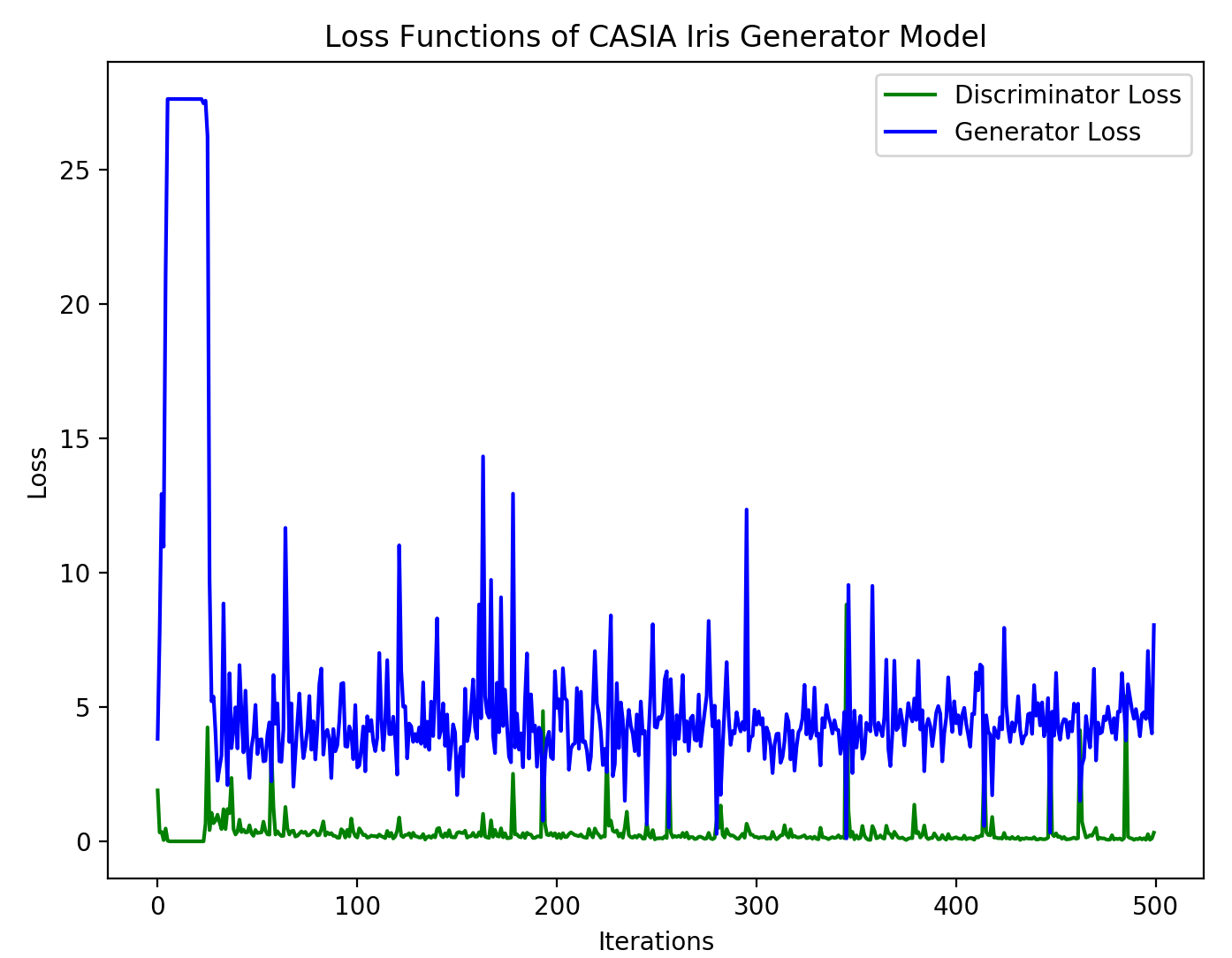}
\end{center}
   \caption{The discriminator and generator loss functions over different iterations, for the CASIA model.}
\label{fig:casialoss}
\end{figure}

\begin{figure}[h]
\begin{center}
   \includegraphics[width=0.95\linewidth]{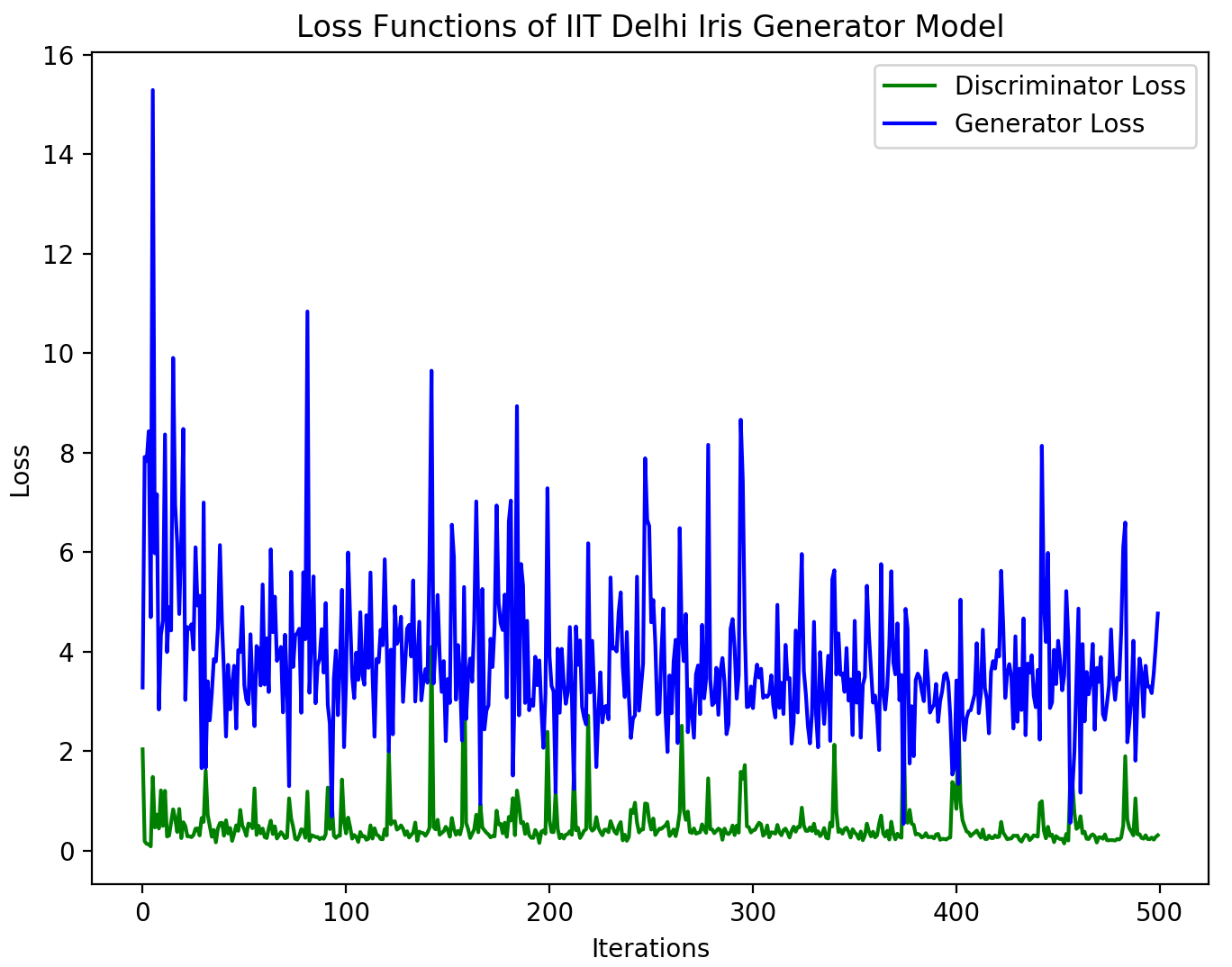}
\end{center}
   \caption{The discriminator and generator loss functions over different iterations for the IIT Delhi model.}
\label{fig:iitdloss}
\end{figure}

We now present the discriminator and generator loss functions for the trained Iris-GAN models on CASIA and IIT Delhi iris databases.
These loss functions for the models trained on CASIA and IIT Delhi databases are presented in Figures \ref{fig:casialoss} and \ref{fig:iitdloss}.
As we can see, it is very hard to interpret the model performance based on the loss values.

\subsection{Numerical Results}
To  measure  the quality  and  diversity of the model numerically,  we computed the Frechet Inception Distance (FID) \cite{FID} on the generated iris images by our model.
FID is an extension of Inception Score (IS) \cite{IS}, which was previously used for assessing the quality of generated images.
FID compares the statistics of generated samples to real samples, using the Frechet distance between two multivariate Gaussians, defined as below:
\begin{equation}
\begin{aligned}
& \textbf{FID}=  \|\mu_r - \mu_g \|^2 + \textbf{Tr} (\Sigma_r + \Sigma_g - 2(\Sigma_r \Sigma_g)^{(1/2)} )
\end{aligned}
\end{equation}
where $X_r \sim N(\mu_r, \Sigma_r)$ and $X_g \sim N(\mu_g, \Sigma_g)$ are the 2048-dimensional activations of the Inception-v3 pool3 layer for real and generated samples respectively.

The FID scores by our proposed model on both CASIA-Iris and IIT Delhi iris databases are shown in Table 1.
As we can see, the model achieves relatively low FID score, comparable with some of state-of-the-arts models on public image datasets.

\begin{table}[ht]
\centering
  \caption{Frechet Inception Distance achieved by our models trained on CASIA-1000, and IIT-Delhi iris databases.}
  \centering
\begin{tabular}{|m{3cm}|m{1.5cm}|}
\hline
Model/Database  & \ \ FID score\\
\hline
DC-GAN/ CASIA-1000  & \ \ \ \ \  42.1\%\\
 \hline 
DC-GAN/ IIT-Delhi &   \ \ \ \ \  41.08\% \\
\hline
\end{tabular}
\label{TblComp}
\end{table}

\section{Conclusion}
\label{sec:Conclusion}
In this work we propose a framework for iris image generation using deep convolutional generative adversarial networks. 
The generator and discriminator networks of our model contain 5 layers (each followed by batch normalization and non-linearity). 
These models are trained on two popular iris databases, CASIA-1000 and IIT-Delhi, and the generator network is used to generate artificial iris images.
The experimental results show that the generated iris images, look very similar to real iris images in those databases, and it is very hard to distinguish them.
In future, we try to extend this framework to subject specific iris image generation, as well as alternative generative models such as variational auto-encoder.

\section*{Acknowledgment}
We would like to thank Center for Biometrics and Security Research (CASIA), IIT Delhi, and the Hong Kong Polytechnic University (PolyU) for providing us the iris databases used in this work.
We would also like to thank Ian Goodfellow for his comments and suggestions regarding numerical analysis.

\end{document}